\pdfoutput=1
\documentclass[letterpaper]{article}
\usepackage[preprint]{aaai2027}
\usepackage[hyphens]{url}
\usepackage{graphicx}
\usepackage{natbib}
\usepackage{caption}
\usepackage{amsmath}
\usepackage{amssymb}
\usepackage{booktabs}
\usepackage{multirow}
\usepackage{array}
\graphicspath{{figures/}}

\title{CoMPASS: Collaborative Molecular Property Prediction via Adaptive Small--Large Model Synergy}
\author{
Wentao Li, Jiangjie Qiu, Yijun Li, Leyi Zhao, Xiaonan Wang\corresponding
}
\affiliations{
Beijing Key Laboratory of Artificial Intelligence for Advanced Chemical Engineering Materials\\
State Key Laboratory of Chemical Engineering and Low-Carbon Technology\\
Department of Chemical Engineering, Tsinghua University\\
Beijing 100084, China
}

\begin{document}

\maketitle

\begin{abstract}
Accurate molecular property prediction requires both statistical reliability and chemical reasoning.  Graph neural networks can be calibrated directly on labeled assays but remain limited by the coverage of their training data.  Large language models (LLMs) can compare molecular evidence and articulate chemical rationales, yet are unreliable as standalone quantitative predictors.  The central challenge is therefore to determine when an LLM should influence a calibrated model and by how much.  Here we present \textbf{CoMPASS}, a retrieval-calibrated framework for small--large model collaboration.  CoMPASS retains a graph attention network (GAT) as the predictive anchor, retrieves locally relevant training molecules, provides attention-grounded evidence to an LLM, and converts its proposal into a bounded correction through an agreement-aware gate.  Across six classification and two regression benchmarks, CoMPASS improves the GAT anchor in regions of correctable uncertainty while limiting LLM intervention in high-confidence regimes.  Ablations show that the gains arise from validation-calibrated retrieval and bounded fusion rather than prompting alone.  These results suggest that generative reasoning should augment calibrated prediction through evidence-grounded, controlled corrections rather than direct output replacement.  Code is available at \url{https://github.com/littlepeachs/CoMPASS}.
\end{abstract}

\section{Introduction}

Molecular property prediction is useful only when its numerical outputs are reliable enough to guide downstream decisions.  In drug discovery and toxicology, a model must assign a probability or continuous property value that can support screening, prioritization, and risk assessment~\citep{cherkasov2014qsar,vamathevan2019applications,wu2018moleculenet,stokes2020deep}.  This requirement creates a dual demand: the predictor must be statistically calibrated on labeled assays, but it should also reason over chemical evidence when local data are sparse or ambiguous.

Graph neural networks (GNNs), including graph attention networks (GATs), address the supervised prediction side of this problem by learning directly from molecular graphs and task-specific labels~\citep{duvenaud2015convolutional,kearnes2016molecular,gilmer2017neural,kipf2017semi,velickovic2018graph,schutt2018schnet,klicpera2020directional,xiong2020pushing}.  Their reliability, however, is bounded by the coverage of the training distribution.  Near decision boundaries, rare scaffolds, sparse activity regions, or noisy multi-task assays, a compact graph model may return a confident prediction even when the local evidence is weak or conflicting.

LLMs provide a complementary form of evidence processing.  Recent scientific and chemistry-oriented LLMs can follow structured instructions, compare molecular descriptions, and articulate rationales~\citep{taylor2022galactica,bran2023chemcrow,christofidellis2023unifying,grattafiori2024llama3,gemma2024gemma2,qwen2025qwen25,zhao2025chemdfm,zhang2024chemllm}.  Yet a fluent chemical rationale is not the same as a calibrated quantitative prediction.  If an LLM is allowed to freely replace the output of a supervised predictor, retrieved molecules become anecdotes rather than statistical evidence, and the final score can drift away from the assay-calibrated model.

The central challenge is therefore to decide when an LLM should influence a calibrated molecular predictor and by how much.  CoMPASS answers this question by treating large-model reasoning as a controlled correction rather than direct output replacement.  The GAT remains the predictive anchor.  Retrieval supplies locally relevant training molecules and a validation-calibrated numerical estimate.  The LLM receives a structured packet containing the target molecule, GAT prediction, attention evidence, retrieved analogues, and conflict summaries.  Its proposal is then converted into a bounded prediction shift through a deterministic agreement-aware gate.  Thus every LLM-mediated prediction can be traced to four explicit signals: the small model, the retrieved neighborhood, the LLM proposal, and the gate that mediates the correction.

\textbf{Contributions.}  (1) We formulate small--large collaboration for molecular property prediction as evidence-grounded correction of a calibrated graph predictor.  (2) We introduce retrieval-calibrated base prediction, where retrieved neighbors serve both as prompt evidence and validation-fitted numerical calibration.  (3) We define an agreement-aware gate that limits LLM intervention in high-confidence regimes while allowing correction in uncertain regions.  (4) We validate the mechanism across classification and regression benchmarks using ablations, confidence-bin analysis, backbone swaps, and success/failure cases.

\begin{figure*}[t]
\centering
\includegraphics[width=0.96\textwidth]{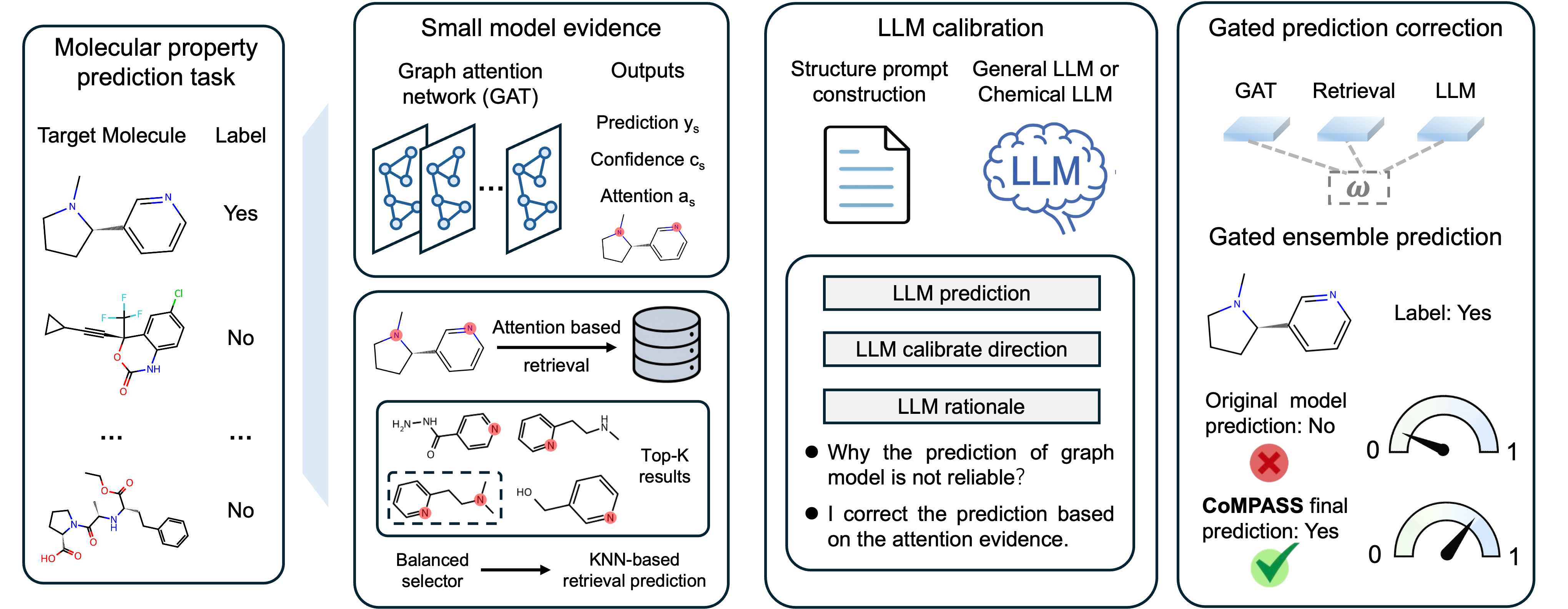}
\caption{\textbf{CoMPASS overview.} A GAT produces the base prediction and attention evidence. Retrieval supplies top-$K$ calibration molecules. The LLM proposes an evidence-conditioned correction, and a bounded gate controls how much that proposal changes the final prediction.}
\label{fig:method}
\end{figure*}

\section{Method}

\subsection{Problem Setup and Small-Model Evidence}

Given a molecule represented by a SMILES string $s$ and graph $G$, the goal is to predict a property $y$.  Classification tasks are evaluated by ROC-AUC, and regression tasks by RMSE.  CoMPASS decomposes prediction into four stages.  First, a GAT produces a supervised prediction, confidence, attention evidence, and graph representation:
\begin{equation}
(\hat{y}_s,c_s,a_s,h_s)=f_s(G).
\end{equation}
For classification, confidence is prediction sharpness,
\begin{equation}
c_s^{\mathrm{cls}}=\max(\hat{y}_s,1-\hat{y}_s).
\end{equation}
For regression, confidence is derived from Monte Carlo dropout uncertainty~\citep{gal2016dropout},
\begin{equation}
c_s^{\mathrm{reg}}=\frac{1}{1+u_s},
\end{equation}
where $u_s$ is predictive standard deviation on the standardized target scale.  Attention is not used only for visualization: it becomes a communication channel that tells the LLM which atoms or functional groups the graph model considered important.

\subsection{Retrieval-Calibrated Base Prediction}

For each target molecule, the retriever returns $K$ calibration molecules from the training split only:
\begin{equation}
\mathcal{R}_K(s)=
\operatorname{TopK}_{i\in\mathcal{D}_{\mathrm{train}}}
\mathrm{sim}(s,s_i).
\end{equation}
Similarity combines Morgan fingerprint overlap, graph-embedding cosine similarity, and attention overlap:
\begin{equation}
\mathrm{sim}(i,j)=\omega_m m_{ij}+\omega_q q_{ij}+\omega_a r_{ij}.
\end{equation}
The query molecule is excluded by SMILES match, and validation/test molecules are never inserted into the retrieval database.  Retrieved labels form a similarity-weighted local estimate:
\begin{align}
\tilde{w}_i &= \max(\mathrm{sim}(s,s_i),0)+\epsilon,\\
w_i &= \tilde{w}_i \Big/ \sum_{j\in\mathcal{R}_K}\tilde{w}_j,\\
\hat{y}_r &= \sum_{i\in\mathcal{R}_K(s)} w_i y_i .
\end{align}
For classification, $\hat{y}_r$ is a probability-like label aggregate; for regression, it is a local property estimate on the standardized target scale.

CoMPASS forms a retrieval-calibrated base prediction,
\begin{equation}
\hat{y}_b=(1-\lambda)\hat{y}_s+\lambda\hat{y}_r,
\end{equation}
where $\lambda$ is selected on validation data for each dataset and retrieval size:
\begin{equation}
\lambda^\star=\arg\min_{\lambda\in\Lambda}
\mathcal{L}_{\mathrm{val}}\big((1-\lambda)\hat{y}_s+\lambda\hat{y}_r,y\big).
\end{equation}
For classification, $\mathcal{L}_{\mathrm{val}}$ is negative ROC-AUC; for regression, it is RMSE.  This makes retrieval more than prompt context: local examples also provide a validation-calibrated numerical correction.

\subsection{Structured LLM Proposal}

The prompt presents the property, target SMILES, GAT prediction and confidence, attention-highlighted functional groups, retrieved molecules with labels and GAT predictions, retrieval conflict, attention consistency, and required output format.  The LLM returns
\begin{equation}
(\hat{y}_\ell,c_\ell,d_\ell,g_\ell,r_\ell)=\mathrm{LLM}(P_s),
\end{equation}
where $\hat{y}_\ell$ is the numeric proposal, $d_\ell$ is the correction direction, $g_\ell$ is an optional suggested gate, and $r_\ell$ is the rationale.  The parser uses the last explicit ``Final Prediction'' field to avoid echoed template text.  If parsing fails, CoMPASS falls back to the small-model prediction.  Classification outputs are clipped to $[0,1]$, and regression outputs are clipped to the observed standardized range.

\subsection{Agreement-Aware Bounded Gate}

The final prediction interpolates between the calibrated base and the LLM proposal:
\begin{equation}
\hat{y}=(1-g)\hat{y}_b+g\hat{y}_\ell.
\end{equation}
The gate $g$ is deterministic and bounded.  It uses normalized uncertainty, retrieval conflict, attention consistency, and directional agreement.  Retrieval conflict is
\begin{equation}
\rho=\operatorname{std}_{j\in\mathcal{R}_K(s)}(y_j),
\qquad
\bar{\rho}=\mathrm{clip}(2\tilde{\rho},0,1),
\end{equation}
where $\tilde{\rho}=\rho$ for classification and $\tilde{\rho}=\rho/\sigma_y$ for regression.  Attention consistency is the mean attention-overlap similarity:
\begin{equation}
A_s=\frac{1}{K}\sum_{j\in\mathcal{R}_K}\mathrm{sim}_{\mathrm{att}}(s,s_j).
\end{equation}
Let $\Delta_r=\hat{y}_r-\hat{y}_s$ and $\Delta_\ell=\hat{y}_\ell-\hat{y}_b$.  Directional agreement is
\begin{equation}
\delta_{\ell r}=
\begin{cases}
\mathrm{agree}, & \Delta_r\Delta_\ell>0,\\
\mathrm{weak}, & |\Delta_r|<\xi\ \mathrm{or}\ |\Delta_\ell|<\xi,\\
\mathrm{disagree}, & \mathrm{otherwise}.
\end{cases}
\end{equation}
The implemented gate is
\begin{equation}
g=\mathrm{clip}\!\left(
s_{\delta_{\ell r}}\Phi_\Theta(g_\ell,\mathbf{e}_s),
f_{\delta_{\ell r}},c_{\delta_{\ell r}}
\right),
\end{equation}
where
\begin{equation}
\mathbf{e}_s=[1,1-c_s,\bar{u}_s,\bar{\rho},1-A_s]^\top.
\end{equation}
The normalized uncertainty term is $\bar{u}_s=\mathrm{clip}(u_s/\log 2,0,1)$ for classification entropy and $\bar{u}_s=\mathrm{clip}(u_s,0,1)$ for standardized regression dropout uncertainty.
Here $\Phi_\Theta$ is a fixed linear evidence map blended with the optional LLM-recommended gate.  The floor $f_\delta$, cap $c_\delta$, and scale $s_\delta$ depend on agreement: agreeing retrieval and LLM shifts receive the largest cap, weak agreement receives a smaller cap, and disagreement receives the most conservative cap.  Thus the LLM always participates through a capped correction rather than direct label replacement.

\begin{table}[t]
\centering
\setlength{\tabcolsep}{2.3pt}
\begin{tabular}{@{}p{0.97\columnwidth}@{}}
\toprule
\textbf{CoMPASS inference for one test molecule} \\
\midrule
1. Run the GAT: obtain $\hat{y}_s,c_s,a_s,h_s$. \\
2. Retrieve $\mathcal{R}_K(s)$ from the training split and compute $\hat{y}_r$. \\
3. Form $\hat{y}_b=(1-\lambda^\star)\hat{y}_s+\lambda^\star\hat{y}_r$. \\
4. Prompt the LLM with target, GAT evidence, and retrieved cases. \\
5. Parse $\hat{y}_\ell,g_\ell,r_\ell$; fall back to $\hat{y}_s$ if parsing fails. \\
6. Compute agreement $\delta_{\ell r}$ and bounded gate $g$. \\
7. Return $\hat{y}=(1-g)\hat{y}_b+g\hat{y}_\ell$ and the full evidence trace. \\
\bottomrule
\end{tabular}
\caption{\textbf{Inference summary.} CoMPASS makes every LLM-mediated prediction auditable through the GAT output, retrieved cases, LLM proposal, and gate.}
\label{tab:algorithm}
\end{table}

\section{Experimental Setup}

\subsection{Datasets, Metrics, and Splits}

We evaluate eight benchmarks spanning target activity, pharmacokinetics, toxicity, and physicochemical regression.  BACE, BBBP, ClinTox, HIV, Tox21, ESOL, and Lipo are standard MoleculeNet-style benchmarks~\citep{wu2018moleculenet}; CYP450 is a five-task cytochrome P450 inhibition benchmark commonly used for ADME modeling~\citep{huang2021therapeutics}.  Six tasks use ROC-AUC; ESOL and Lipo use original-scale RMSE.  All runs use fixed DeepChem train/validation/test splits.  For multi-task datasets, predictions, labels, and weights are stored as task vectors; retrieval produces a label vector, validation fitting respects task weights, and final metrics average valid task entries.

\begin{table*}[t]
\centering
\setlength{\tabcolsep}{4.0pt}
\begin{tabular}{@{}llclll@{}}
\toprule
Dataset & Type & Tasks & Train/Valid/Test & Label statistic & Metric \\
\midrule
BACE & Class. & 1 & 1210/151/152 & 0.426 positive & ROC-AUC \\
BBBP & Class. & 1 & 1631/204/204 & 0.822 positive & ROC-AUC \\
ClinTox & Class. & 2 & 1184/148/148 & 0.507 positive & ROC-AUC \\
CYP450 & Class. & 5 & 13040/1690/1690 & 0.305 positive & ROC-AUC \\
HIV & Class. & 1 & 32896/4112/4112 & 0.037 positive & ROC-AUC \\
Tox21 & Class. & 12 & 6258/782/783 & 0.071 positive & ROC-AUC \\
ESOL & Reg. & 1 & 902/113/113 & -2.867 train mean & RMSE \\
Lipo & Reg. & 1 & 3360/420/420 & 2.163 train mean & RMSE \\
\bottomrule
\end{tabular}
\caption{\textbf{Dataset statistics.} Counts are fixed DeepChem splits. Classification statistics are positive rates. Regression statistics are train-label means on the original property scale.}
\label{tab:datasets}
\end{table*}

\subsection{Configurations and Selection Protocol}

The main configuration uses Llama-3-8B and evaluates $K\in\{4,6,8\}$.  Retrieval weight $\lambda^\star$ and the reported retrieval size are selected on validation data.  Test labels are used only for final evaluation.  The main results average three independent seeds, with GAT training for up to 30 epochs with early stopping, inference batch size 32, learning rate $10^{-4}$, weight decay $10^{-5}$, Morgan radius 2 with 2048 bits, and LLM decoding temperature 0.7.  Additional experiments compare Qwen2.5-7B, ChemDFM, and ChemLLM, and replace the GAT anchor with a GraphGPS-style graph transformer~\citep{rampasek2022recipe} to test whether the collaboration interface is tied to one graph architecture.

\section{Results}

\subsection{Main Results: Recoverable Anchor Errors}

\begin{table*}[t]
\centering
\setlength{\tabcolsep}{3.0pt}
\begin{tabular*}{\textwidth}{@{\extracolsep{\fill}}llcccccc@{}}
\toprule
Model & Method &
\shortstack{BACE\\(152)} &
\shortstack{BBBP\\(204)} &
\shortstack{ClinTox\\(148)} &
\shortstack{CYP450\\(1690)} &
\shortstack{HIV\\(4112)} &
\shortstack{Tox21\\(783)} \\
\midrule
\multirow{5}{*}{\shortstack{Pre-training\\Methods}}
& Gimlet & 0.696 & 0.594 & -- & 0.713 & 0.662 & 0.612 \\
& KVPLM & 0.513 & 0.602 & -- & 0.592 & 0.612 & 0.492 \\
& MoMu & 0.666 & 0.498 & -- & 0.580 & 0.503 & 0.576 \\
& Galactica-125M & 0.445 & 0.605 & -- & 0.537 & 0.367 & 0.496 \\
& Galactica-1.3B & 0.565 & 0.539 & -- & 0.469 & 0.339 & 0.495 \\
\midrule
\multirow{2}{*}{MolRAG}
& Struct-CoT & 0.626 & 0.572 & -- & 0.584 & 0.595 & 0.566 \\
& Sim-CoT & 0.723 & 0.541 & -- & 0.723 & 0.644 & 0.639 \\
\midrule
\multirow{3}{*}{\shortstack{Graph-based\\Networks}}
& GAT & 0.756 & 0.652 & \underline{0.893} & \underline{0.877} & 0.715 & 0.735 \\
& GIN & 0.701 & 0.658 & -- & 0.821 & \underline{0.753} & 0.740 \\
& Graphormer & \underline{0.776} & \underline{0.702} & -- & 0.844 & 0.745 & \textbf{0.759} \\
\midrule
\multirow{2}{*}{Our runs}
& \textbf{CoMPASS} &
\textbf{0.852} &
\textbf{0.732} &
\textbf{0.937} &
\textbf{0.880} &
\textbf{0.755} &
\underline{0.744} \\
& Gain vs GAT & +0.096 & +0.080 & +0.044 & +0.003 & +0.040 & +0.009 \\
\bottomrule
\end{tabular*}
\caption{\textbf{Experimental results on classification test datasets.} Values are ROC-AUC, where higher is better. External GIMLET, KVPLM, MoMu, Galactica, GIN, and Graphormer values are reproduced from \citet{zhao2023gimlet}; four-shot MolRAG Struct-CoT and Sim-CoT values are reproduced from \citet{xian2025molrag}. Only GAT and CoMPASS are paired under the fixed split and evaluator. The best value in each dataset column is bolded and the second-best value is underlined.}
\label{tab:main-results}
\end{table*}

\begin{table}[t]
\centering
\setlength{\tabcolsep}{3pt}
\begin{tabular*}{\columnwidth}{@{\extracolsep{\fill}}llcc@{}}
\toprule
Model & Method &
\shortstack{ESOL\\RMSE $\downarrow$} &
\shortstack{Lipo\\RMSE $\downarrow$} \\
\midrule
\multirow{3}{*}{Llama3-8B}
& 1-shot Sim-CoT & 4.142 & 1.271 \\
& 2-shot Sim-CoT & 3.499 & 1.168 \\
& 4-shot Sim-CoT & 3.281 & 1.125 \\
\midrule
\shortstack{Pre-train.\\Methods} & Gimlet & \textbf{1.132} & 1.345 \\
\midrule
\multirow{2}{*}{\shortstack{Graph\\Networks}}
& GAT & 1.275 & 0.816 \\
& GIN & 1.243 & \textbf{0.781} \\
\midrule
\multirow{2}{*}{Our runs}
& \textbf{CoMPASS} & \underline{1.201} & \underline{0.790} \\
& RMSE reduction vs GAT & +0.074 & +0.026 \\
\bottomrule
\end{tabular*}
\caption{\textbf{Regression test results.} Values are original-scale RMSE, so lower is better. Llama3-8B Sim-CoT values at 1, 2, and 4 shots are reproduced from \citet{xian2025molrag}, and external Gimlet and GIN values are reproduced from \citet{zhao2023gimlet}; only GAT and CoMPASS are paired under the fixed split and evaluator. The best value in each dataset column is bolded and the second-best value is underlined.}
\label{tab:regression-results}
\end{table}

Table~\ref{tab:main-results} and Table~\ref{tab:regression-results} show that CoMPASS consistently improves the paired GAT anchor, but it does not claim universal dominance over all graph baselines. Its value is clearest where the anchor leaves retrievable local error. BACE, BBBP, ClinTox, and HIV show large improvements because retrieved analogues expose boundary or calibration mistakes. CYP450 and Tox21 have smaller gains because the GAT anchor is already strong and the gate mainly limits harm. Regression follows the same pattern: CoMPASS improves over our GAT runs on ESOL and Lipo, while the claim remains calibrated correction rather than state-of-the-art regression.

\begin{table}[t]
\centering
\setlength{\tabcolsep}{4.0pt}
\begin{tabular}{@{}lcc@{}}
\toprule
Dataset & Gain & Bootstrap interval \\
\midrule
BACE & +0.096 & [0.038, 0.155] \\
BBBP & +0.080 & [0.041, 0.121] \\
ClinTox & +0.044 & [0.018, 0.081] \\
CYP450 & +0.003 & [-0.000, 0.006] \\
HIV & +0.040 & [0.022, 0.060] \\
Tox21 & +0.009 & [0.001, 0.016] \\
ESOL & +0.074 & [0.018, 0.134] \\
Lipo & +0.026 & [0.013, 0.040] \\
\bottomrule
\end{tabular}
\caption{\textbf{Molecule-bootstrap gain intervals.} Positive gain means ROC-AUC increase for classification and original-scale RMSE reduction for regression.}
\label{tab:bootstrap}
\end{table}

Bootstrap intervals in Table~\ref{tab:bootstrap} are positive on BACE, BBBP, ClinTox, HIV, ESOL, and Lipo.  Tox21 is positive but small, and CYP450 is best read as a low-headroom setting where bounded gating preserves the anchor.  This supports the central claim: CoMPASS helps when local evidence reveals correctable GAT uncertainty, and its conservative design matters when headroom is limited.

\subsection{Retrieval Size Is a Calibration Axis}

\begin{figure*}[t]
\centering
\includegraphics[width=0.96\textwidth]{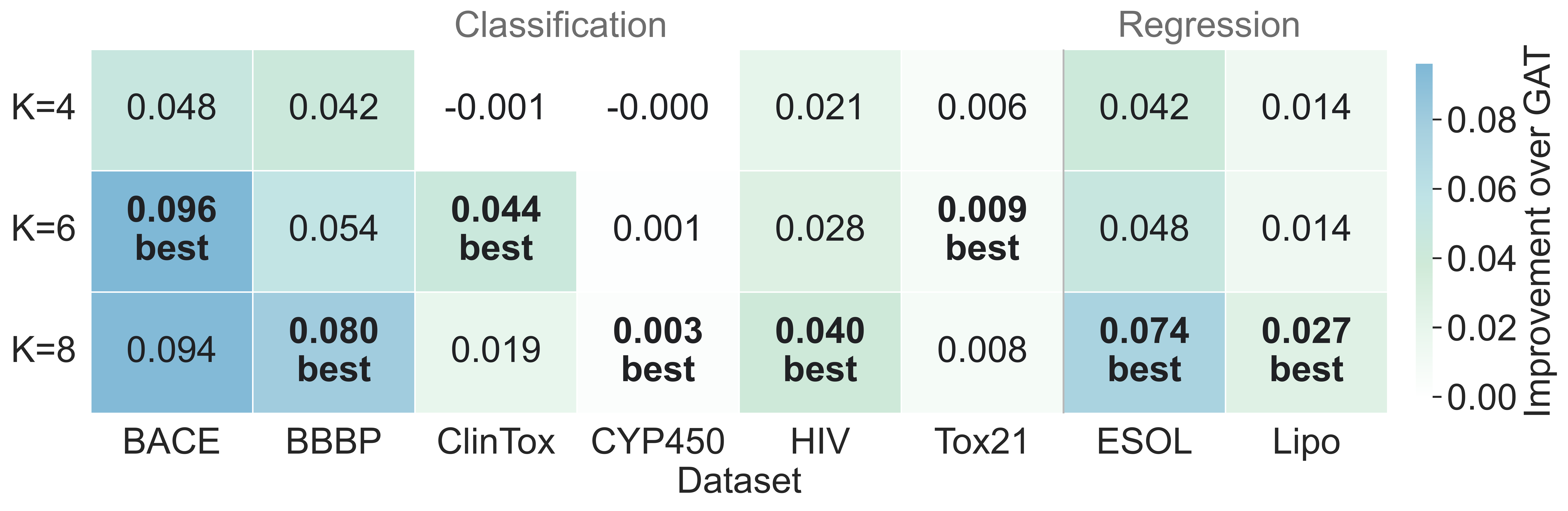}
\caption{\textbf{Main-result heatmap over retrieval size.} Cells show improvement over the GAT baseline in each dataset's metric units: ROC-AUC points for classification and original-scale RMSE reduction for regression.}
\label{fig:k-sensitivity}
\end{figure*}

Figure~\ref{fig:k-sensitivity} shows that retrieval size is not a cosmetic hyperparameter.  BACE, ClinTox, and Tox21 peak at $K=6$, while BBBP, HIV, ESOL, and Lipo peak at $K=8$.  More neighbors help when analogues form a coherent local neighborhood, but can dilute signal when the property varies sharply.  CoMPASS therefore treats $K$ as a validation-calibrated design axis: it determines whether the LLM sees a focused local counterfactual set or a noisier neighborhood summary.

\subsection{Ablations: Retrieval Carries Signal, Gating Controls It}

\begin{table*}[t]
\centering
\setlength{\tabcolsep}{4.0pt}
\begin{tabular*}{\textwidth}{@{\extracolsep{\fill}}lrrrr@{}}
\toprule
Variant & BACE & BBBP & ESOL & Lipo \\
\midrule
GAT only & 0.756 & 0.652 & 1.275 & 0.816 \\
LLM + GAT & 0.759 & 0.655 & 1.247 & 0.818 \\
GAT + Attention Retrieval & 0.781 & 0.663 & 1.244 & \underline{0.802} \\
GAT + Similarity Retrieval & 0.796 & 0.662 & 1.226 & 0.809 \\
LLM + GAT + Retrieval (fixed weight) & \underline{0.819} & \underline{0.687} & \underline{1.216} & 0.804 \\
LLM + GAT + Retrieval (no attention) & 0.781 & 0.661 & 1.230 & 0.812 \\
\textbf{Full CoMPASS} & \textbf{0.852} & \textbf{0.732} & \textbf{1.201} & \textbf{0.790} \\
\bottomrule
\end{tabular*}
\caption{\textbf{Complete ablation study.} BACE and BBBP report ROC-AUC, where higher is better; ESOL and Lipo report original-scale RMSE, where lower is better. The best value in each dataset column is bolded and the second-best value is underlined.}
\label{tab:ablation}
\end{table*}

Table~\ref{tab:ablation} separates three effects.  First, LLM-only correction without calibrated retrieval is weak, so the LLM should not replace the GAT.  Second, retrieval carries meaningful signal, but fixed or partial retrieval variants are weaker than the full interface.  Third, attention evidence helps ground the LLM when retrieved molecules contain mixed evidence.  The useful signal is therefore not the raw LLM estimate alone; it is the agreement-filtered correction after retrieval calibration.

This ablation also clarifies the role of the gate.  Retrieval-only variants can already improve regression and some classification tasks, which means local label evidence is genuinely informative.  However, prompt evidence and LLM rationales become useful only when the final score is capped by agreement and uncertainty.  CoMPASS should therefore be viewed as a calibrated decision rule for incorporating generative reasoning, not as a prompt engineering method.

\subsection{Backbone and Anchor Robustness}

\begin{table}[t]
\centering
\setlength{\tabcolsep}{3.0pt}
\begin{tabular}{@{}lccccc@{}}
\toprule
Dataset & GAT & L3 & Qwen & CDFM & CLLM \\
\midrule
BACE & 0.756 & \textbf{0.852} & 0.849 & 0.843 & 0.848 \\
BBBP & 0.652 & \textbf{0.732} & 0.717 & 0.701 & 0.707 \\
ESOL & 1.275 & 1.201 & 1.231 & \textbf{1.183} & 1.210 \\
Lipo & 0.816 & 0.790 & 0.798 & 0.800 & \textbf{0.789} \\
\bottomrule
\end{tabular}
\caption{\textbf{LLM backbone comparison.} BACE/BBBP use ROC-AUC; ESOL/Lipo use RMSE. L3 denotes Llama-3-8B, CDFM denotes ChemDFM, and CLLM denotes ChemLLM.}
\label{tab:backbone}
\end{table}

\begin{table}[t]
\centering
\setlength{\tabcolsep}{4.0pt}
\begin{tabular}{@{}llccc@{}}
\toprule
Data & Metric & GPS & CoMPASS-GPS & Gain \\
\midrule
BACE & ROC & 0.770 & \textbf{0.868} & +0.098 \\
BBBP & ROC & 0.718 & \textbf{0.730} & +0.012 \\
ESOL & RMSE & 1.107 & \textbf{1.103} & +0.005 \\
\bottomrule
\end{tabular}
\caption{\textbf{GraphGPS anchor swap.} Only the graph anchor is changed; retrieval, prompting, and gating remain unchanged.}
\label{tab:gps}
\end{table}

Table~\ref{tab:backbone} shows that the collaboration interface is not tied to one LLM.  Llama-3-8B is strongest on BACE and BBBP, while chemistry-oriented LLMs are competitive on regression.  This suggests a practical deployment path: the same GAT and retriever can route difficult examples to different LLMs according to endpoint type or local availability.  Table~\ref{tab:gps} shows that the interface can also plug into an attention-bearing graph transformer.  The effect is largest on BACE and smaller on BBBP/ESOL, consistent with the idea that collaboration helps most when the anchor leaves retrievable headroom.

\subsection{Where Corrections Help}

\begin{figure}[t]
\centering
\includegraphics[width=\columnwidth]{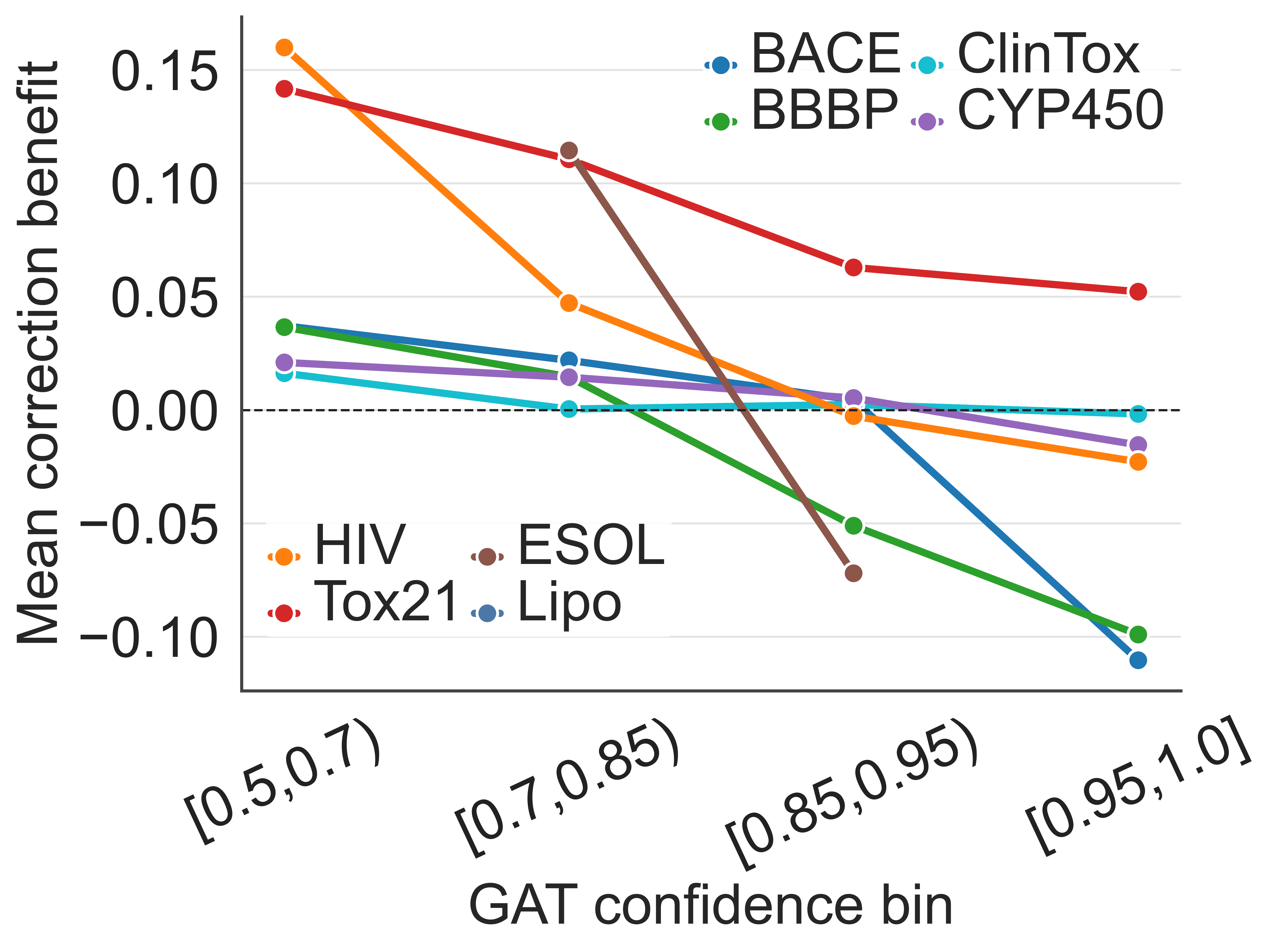}
\caption{\textbf{Confidence-binned behavior.} Mean correction benefit is largest in lower-confidence GAT bins and shrinks as the GAT becomes more confident.}
\label{fig:confidence}
\end{figure}

The confidence-binned analysis in Figure~\ref{fig:confidence} connects the metric gains to the gate design.  Low-confidence GAT predictions have more headroom and benefit most from retrieved evidence and LLM interpretation.  High-confidence predictions have less headroom and greater over-correction risk.  This pattern explains why an unrestricted LLM mixture is not desirable: the large model should be heard most clearly when the anchor is uncertain, and should be capped when the anchor is confident or retrieval evidence is inconsistent.

\section{Case and Failure Analysis}

\begin{figure*}[t]
\centering
\includegraphics[width=0.77\textwidth]{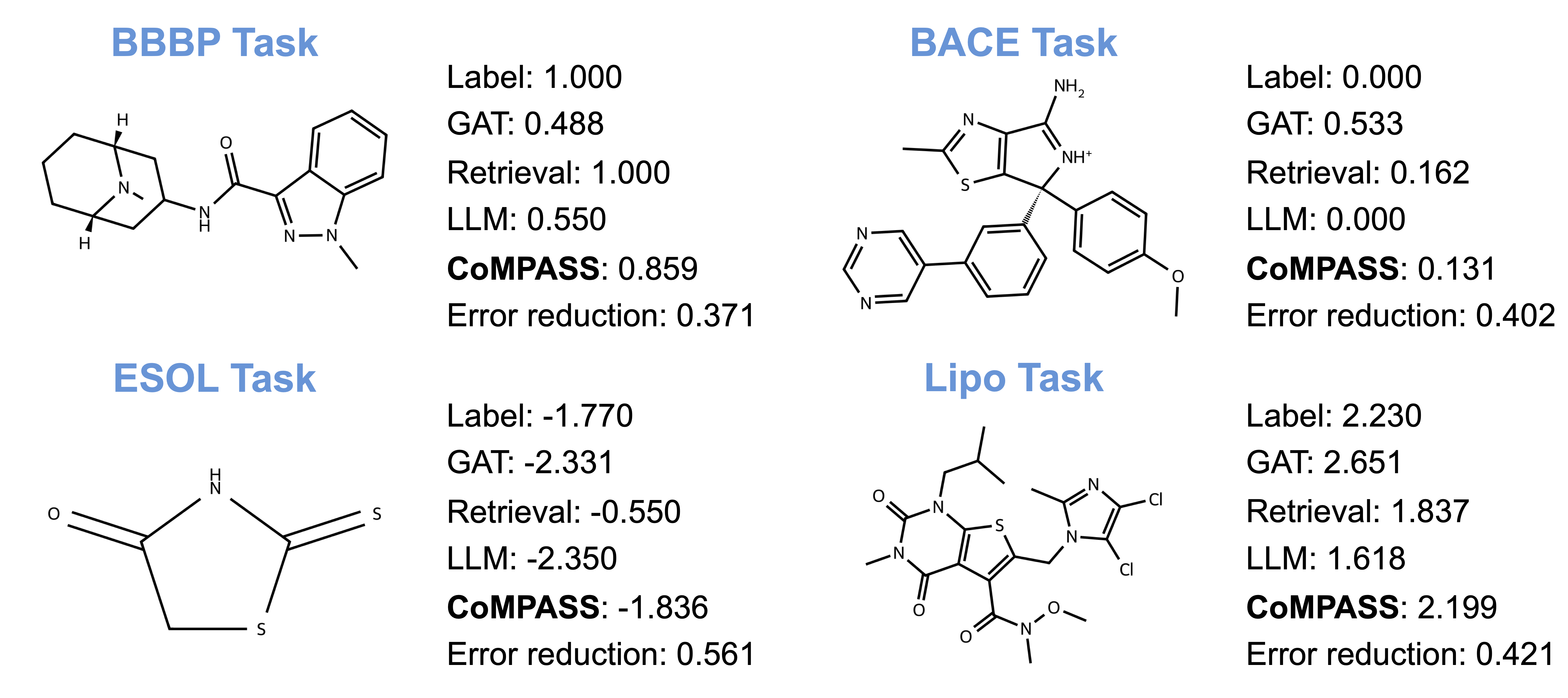}
\includegraphics[width=0.77\textwidth]{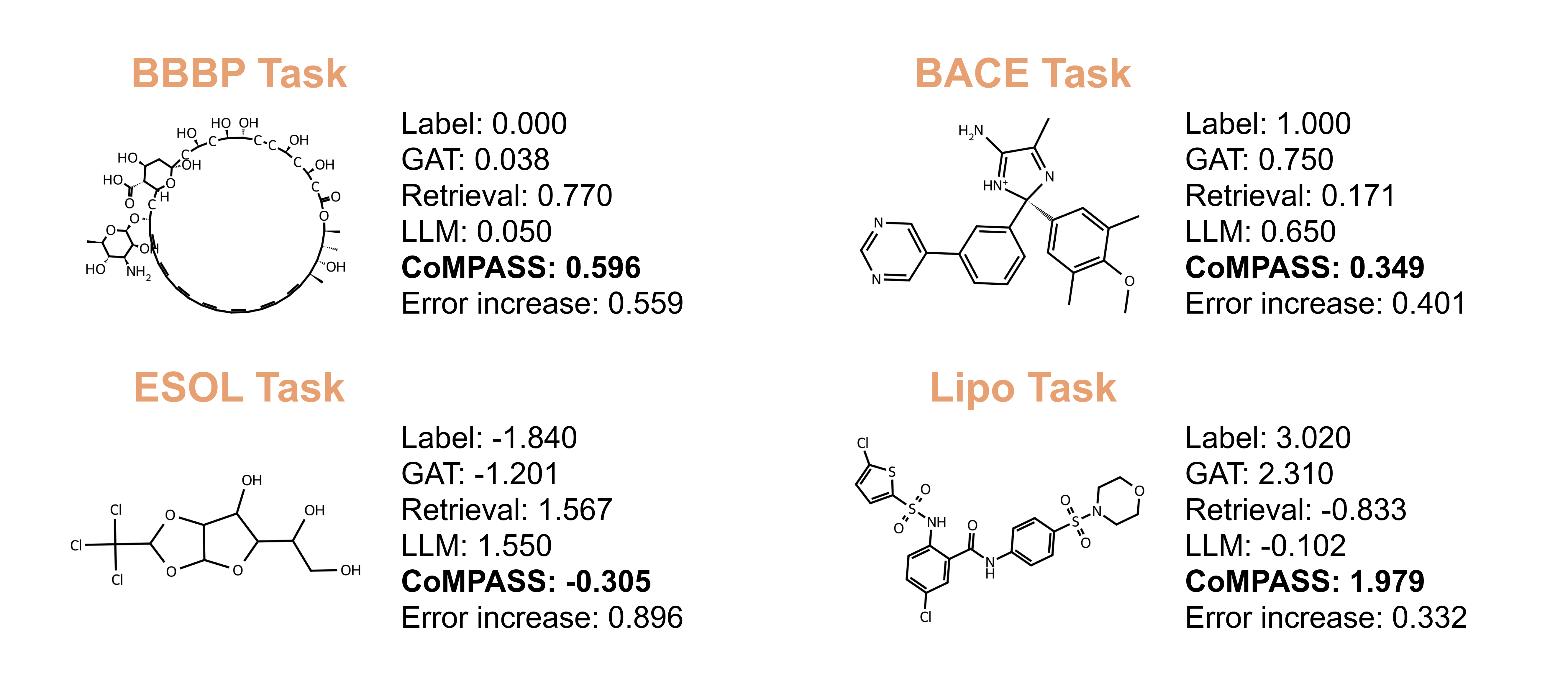}
\caption{\textbf{Molecule-level success and failure cases.} Top: successful corrections show boundary movement or regression error reduction. Bottom: failure cases show over-correction from misleading retrieval or a nonzero gate floor.}
\label{fig:case-failure}
\end{figure*}

Figure~\ref{fig:case-failure} illustrates the two correction regimes.  In classification, successful cases often involve boundary movement: the GAT score lies near the 0.5 threshold, retrieved analogues provide a directional signal, and the gate permits enough movement to change the decision.  In regression, there is no discrete threshold; the correction usually moves partway toward a local physicochemical property scale.  The point is not that the LLM invents a label, but that it interprets structured evidence and proposes a direction that survives bounded fusion.

The lower half of Figure~\ref{fig:case-failure} reveals the main limitation.  Retrieval can be misleading: a correct GAT prediction can be pulled across the threshold, or a regression estimate can move farther from the label.  The nonzero gate floor is intentional because a zero floor would make the large model decorative, but the cap is equally important because it prevents uncontrolled override.  Future gains are therefore more likely to come from better conflict detection and more adaptive caps than from simply increasing the LLM weight.  Showing both successful and failed corrections is central to the CoMPASS claim: the method is not a promise that LLMs always help, but a controlled interface that makes their contribution and risk auditable.

\section{Operational Interpretation}

\textbf{Multi-task endpoints.}  ClinTox, CYP450, and Tox21 contain multiple assay tasks with missing-label masks.  CoMPASS keeps the GAT prediction as a task vector and computes retrieval-calibrated label aggregates with the same evaluator masks used by the benchmark.  The LLM emits a scalar correction because the prompt asks it to reason over the molecule-level evidence packet rather than produce one free-form answer per assay.  This scalar shift is applied to the retrieval-calibrated task vector and clipped to the valid range.  The gate is therefore molecule-specific rather than task-specific, while metric computation remains task-aware through the original masks.  This design avoids asking the LLM to hallucinate labels for missing tasks and keeps the supervised graph model responsible for task-specific calibration.

\textbf{Audit trail.}  Each CoMPASS prediction stores the target SMILES, GAT score, confidence, attention summary, retrieved molecules, retrieved labels, retrieval weights, LLM proposal, gate value, final prediction, and rationale.  This trace is not just a logging convenience.  It allows a reviewer or practitioner to distinguish three cases that have the same final metric effect but different operational meaning: a retrieval-driven correction, an LLM proposal suppressed by disagreement, and an LLM proposal accepted because it agrees with local evidence.  The trace also enables the failure gallery in Figure~\ref{fig:case-failure}; without it, a molecular LLM system can look persuasive while hiding whether the large model actually improved the calibrated prediction.

\textbf{Why not prompt-only RAG?}  Prompt-only molecular RAG treats retrieved examples mainly as context for a generator.  CoMPASS instead gives retrieval a numerical role before the LLM is consulted.  This matters because a few retrieved analogues can be chemically plausible but statistically misleading.  Validation-fitted blending estimates how much local labels should move the anchor on each dataset, while the gate determines how much of the LLM's interpretation of that evidence should survive.  The ablation table shows the consequence: prompting and retrieval are useful only when the final prediction remains bounded by calibrated evidence.

\section{Related Work}

\textbf{Molecular representation learning.} Molecular property prediction has long used graph and fingerprint representations~\citep{cherkasov2014qsar,rogers2010extended,wu2018moleculenet,mayr2016deeptox}.  Modern GNNs learn task-specific representations~\citep{duvenaud2015convolutional,kearnes2016molecular,gilmer2017neural,kipf2017semi,velickovic2018graph,xu2019how,ying2021transformers,schutt2018schnet,klicpera2020directional}, while molecular pretraining and contrastive learning improve transfer across endpoints~\citep{hu2020strategies,rong2020self,liu2022pretraining,wang2022molecular,you2020graph}.

\textbf{LLMs and retrieval.} LLMs have been explored for scientific reasoning and chemistry workflows~\citep{taylor2022galactica,bran2023chemcrow,christofidellis2023unifying,zhao2025chemdfm,zhang2024chemllm}.  Retrieval-augmented generation grounds LLM outputs in external examples~\citep{lewis2020retrieval,guu2020realm,karpukhin2020dense,izacard2021leveraging,borgeaud2022retro}.  CoMPASS differs from LLM-driven molecular RAG by making the graph model the primary predictor and using retrieval as both prompt evidence and numeric calibration.

\textbf{Small--large collaboration.} Small--large model collaboration appears in speculative decoding, mixture-of-experts, and adaptation~\citep{leviathan2023fast,fedus2022switch,xu2024small}.  CoMPASS contributes a molecular prediction instance of this paradigm where attention evidence, retrieval calibration, and uncertainty-aware gating mediate the collaboration.  The broader lesson is that large models can improve prediction systems when their role is formalized as bounded, evidence-conditioned correction.

\section{Conclusion}

CoMPASS argues for a restrained role for LLMs in molecular property prediction.  The large model should not replace the supervised graph predictor; it should interpret retrieved analogues and graph-attention evidence when the small model may be locally miscalibrated.  Across eight benchmarks, this interface improves the paired GAT anchor while exposing where headroom is small.  Ablations sharpen the conclusion: retrieval and validation-fitted local correction carry much of the signal, and the LLM becomes useful when its proposal is agreement-filtered and bounded.

\section{Limitations}

CoMPASS adds LLM inference, so it is most appropriate for offline screening, triage, and mechanistic analysis rather than billion-scale virtual-library scoring.  It also depends on retrieval quality: meaningful analogues expose local GAT errors, while misleading neighborhoods can induce over-correction.  Finally, results are reported on fixed benchmark splits rather than repeated resampling protocols, so the conclusions should be read as evidence for the joint retrieval-plus-gating interface, not as a claim that prompting alone solves molecular prediction.

\bibliography{compass_references}

\end{document}